%% file: main.tex
\documentclass[conference]{IEEEtran}
\IEEEoverridecommandlockouts
\usepackage{cite}
\usepackage{amsmath,amssymb,amsfonts}
\usepackage{algorithmic}
\usepackage[ruled,vlined,noend]{algorithm2e}

\SetCommentSty{mycommfont}
\usepackage{graphicx}
\usepackage{textcomp}
\usepackage{xcolor}

\newcommand{\sm}[1]{#1} 
\newcommand{\lr}[1]{#1} 

\begin{document}
\title{Neural Paraphrasing by Automatically Crawled \\ and Aligned Sentence Pairs\\
\thanks{This work was done in the context of the joint research laboratory between QuestIT S.r.l. (\texttt{http://www.quest-it.com}) and the Siena Artificial Intelligence Laboratory (SAILab, \texttt{http://sailab.diism.unisi.it}) of the Department of Information Engineering and Mathematics, University of Siena.}
}
\author{\IEEEauthorblockN{Achille Globo\IEEEauthorrefmark{2}, Antonio Trevisi\IEEEauthorrefmark{2}, Andrea Zugarini\IEEEauthorrefmark{1}, Leonardo Rigutini\IEEEauthorrefmark{2}, Marco Maggini\IEEEauthorrefmark{3}, Stefano Melacci\IEEEauthorrefmark{3}}
\IEEEauthorblockA{\IEEEauthorrefmark{1}\textit{DINFO}, \textit{University of Florence}, Florence, Italy, andrea.zugarini@unifi.it}
\IEEEauthorblockA{\IEEEauthorrefmark{2}\textit{QuestIT S.r.l.}, \textit{The Digital Box S.p.a.}, Siena, Italy, \{globo,trevisi,rigutini\}@quest-it.com}
\IEEEauthorblockA{\IEEEauthorrefmark{3}\textit{DIISM}, \textit{University of Siena}, Siena, Italy, \{maggini,mela\}@diism.unisi.it}
}
\maketitle
\begin{abstract}
\input{01_abstract.tex}  
\end{abstract}
\section{Introduction}
\label{sec:intro}
\input{02_introduction.tex}
\section{Automatic \sm{Dataset Construction}}
\label{sec:data}

In this section we describe the proposed method \sm{to automatically generate} aligned sentence corpora. 
\sm{Our approach is rooted on the basic idea} that different news and blog websites \sm{usually} report the same important facts and events using different styles depending on the author and on the editorial orientation of the publisher of the article (\textit{idiolect}).
The system performs a massive crawling of contents from a list of news and blog websites and \sm{it performs an advanced analysis based on popular Natural Language Processing (NLP) techniques, in order} to enrich the raw text with morphological, syntactic and semantic features. 
\sm{Then, an} HCSSS step is used to retrieve the set of \sm{sentences that are mostly similar to a given set of reference sentences, thus producing a number of candidate paraphrases. \sm{This step can be efficiently performed using search engine technologies.} In the following subsections we report the details of each step of the proposed method. The specific use case on the Italian language, together with the details on the involved software solutions, are reported in the last subsection.}

\subsection{Focused Crawling}
\input{03_crawling.tex}
\subsection{Content Analysis and Preprocessing}
\input{04_preprocessing_nlp.tex}

\subsection{Sentence \sm{Selection and} Alignment}
\input{05_sentence_alignment.tex}

\subsection{Case Study}
\input{06_case_study.tex}
\section{Neural Model}
\label{sec:model}
\input{07_model.tex}
\section{Experiments}
\label{sec:exp}
\input{08_experimental_results.tex}
\section{Conclusions and Future Work}
\label{sec:concl}
\input{09_conclusion_future_works.tex}
\bibliography{biblio}{}
\bibliographystyle{plain}
\end{document}

%% file: 01_abstract.tex
Paraphrasing is the task of re-writing an input text \sm{using other words, without altering the meaning of the original content. Conversational systems can exploit automatic paraphrasing to make the conversation more natural, e.g., talking about a certain topic using different paraphrases in different time instants. }
Recently, \sm{the task of automatically generating paraphrases has been approached in the context of Natural Language Generation (NLG).} 
\sm{While many existing systems simply consist in rule-based models}, the recent success of the Deep Neural Networks \sm{in several NLG tasks naturally suggests the possibility of exploiting such networks for generating paraphrases}.
\sm{However, the main obstacle toward neural-network-based paraphrasing} is the lack of large datasets \sm{with aligned pairs of sentences and paraphrases, that are needed to efficiently train the neural models}.
In this paper we present a method for the automatic generation of \sm{large} aligned corpora, \sm{that} is based on the assumption that news and blog websites \sm{talk about the} same events using different narrative styles.
\sm{We propose} a similarity search procedure with linguistic constraints \sm{that, given a reference sentence, is} able to locate the most similar candidate paraphrases out from millions of indexed sentences.
\sm{The data generation process is evaluated in the case of the Italian language, performing experiments using} pointer-based deep neural architectures.


%% file: 02_introduction.tex

\sm{Humans can easily rephrase a text passage without changing the meaning of the considered text, and they can communicate the same piece of information using different words, e.g., during a conversation. Humans have the cognitive ability to deeply understand the contents that are described in the considered text passage, of the involved events and facts. Machines commonly lack this ability, and designing systems that automatically generate paraphrases of an input text is a challenging task that, recently, has been the subject of a number of studies in the Natural Language Generation (NLG) literature.}
In particular, the task of rewriting text has been classified into three \sm{different categories}  \cite{madnani2010generating}: lexical paraphrasing, \sm{that is based or replacing words with other words} with same meaning \cite{bolshakov2004synonymous}\sm{;} phrasal paraphrasing, when the paraphrase is created acting on fragments with \sm{the} same meaning \cite{ganitkevitch2013ppdb}\sm{;} sentential paraphrasing when the paraphrases is performed \sm{at a sentence level, considering} sentences with \sm{the} same meaning \cite{barzilay2001extracting,barzilay2003learning,dolan2004unsupervised}.

\sm{In this paper we let the machine learn to generate paraphrases of an input text, without enforcing any specific word-level, fragment-level, or sentence-level operations. In particular, we follow the idea that paraphrasing can be approached as a special} Statistical Machine Translation task in which the two languages \sm{involved in the translation, referred to as $L_{input}$ and $L_{target}$, respectively, do} coincide ($L_{input} \equiv L_{target}$) but they are modeled with two different languages models, \sm{one that is about the input text and one that is about the target paraphrase.} 
The success of using Deep Neural Networks in Statistical Machine Translation \cite{kalchbrenner2013recurrent}, in particular sequence-to-sequence architectures \cite{sutskever2014sequence}, suggests that the task of text paraphrasing can be addressed using similar approaches.
Unfortunately, while in the Machine Translation field a large number of datasets with millions of aligned pairs of sentences can be easily found \sm{(where each pair is composed of a sentence from $L_{input}$ and the associated translation in $L_{target}$)}, in the case of paraphrasing  \sm{very large} datasets \sm{of \textit{(sentence, paraphrase)} pairs do not exist, especially in the case of languages different from English.}
The creation of such \sm{``supervised'' (i.e., aligned)} dataset requires \sm{an expensive annotation activity which turns out to be} very difficult to scale to large amounts of data. 

We propose a method for building \sm{a dataset of aligned sentences that can be used to train machine learning models to approach the task of} paraphrasing. 
The resulting dataset consists of a set of pairs of textual sentences\sm{, each of them composed of} an \textit{input} sentence and a \textit{target} paraphrase. 
\sm{Our} method is based on the idea that various news and blog websites normally report the same important facts and events using different \textit{idiolects}. 
An \textit{idiolect} is defined as the individual distinctive and unique use of language, concerning the morpho-syntactic and stylistic features.
This  assumption  allows \sm{us} to  model the problem of dataset building as a High\sm{ly} Constrained Sentence Similarity Search (HCSSS), where the results of a search in a document base must satisfy hard linguistic constraints (morphological, syntactic and semantic constraints) in addition  to word co-occurrences.
The \sm{proposed} method has been evaluated by crawling a large number of articles from Italian newspaper sites and blogs\sm{, thus generating a corpus of pairs that we use to train a deep sequence-to-sequence neural network model. In particular, we focus on a Pointer network \cite{gu2016incorporating,vaswani2017attention,see2017get}) that, given an \textit{input} text, learns to generate a possible \textit{target} paraphrase.}
\sm{We report some preliminary results and we discuss the quality and the limits of the considered neural models in function of the properties of the data generated by the proposed mechanism}. 

\sm{This paper is organized as follows. Section \ref{sec:data} describes the proposed process to automatically crawl and align sentence pairs. Section \ref{sec:model} summarizes the properties of the considered deep neural network model, while Section \ref{sec:exp} reports experimental results and discussions about them. Finally, Section \ref{sec:concl} concludes the paper with some proposals of future work.}

%% file: 03_crawling.tex
\sm{The first stage of the proposed approach is aimed at retrieving} a large number of \sm{articles from the web using a focused crawler.}
\sm{In particular, we propose to exploit a list of newspaper and blog websites, together with a set of topics to limit the contents to specific domains: news, culture, economics, nature, politics, society, sports and technology.} 
\lr{The topic oriented crawling can be driven by groups of  keywords for each domain, so that only articles containing these keywords are retrieved.}
\sm{All the listed} websites can be daily crawled and the contents can be organized either by category (topic) and by date in a such way to ease the subsequent alignment phase. At this step, the considered content consists exclusively of raw text and it is saved for the \sm{following content} analysis phase.

%% file: 04_preprocessing_nlp.tex
\sm{Given the raw text crawled so far, the goal of the next stage} is to recover a large number of pairs of sentences, each of which reports the same facts or events using different narrative methods, from the morphological, syntactical and stylistic point of views.
\sm{In order to} build such pairs, \sm{we propose to apply} a search procedure \sm{that is} bounded by a set of linguistic constraints.

\sm{In particular}, before being indexed, we process the raw text \sm{with} a pipeline of \sm{NLP} techniques \sm{that yields an enriched representation of the text with the identified collocations and named entities, where a collocation is an} aggregation of tokens having a specific meaning if co-occurring together, while a named entity represents a specific entity of the real world, \sm{belonging to a set of predefined types} (people, organization, location, etc.). 
For each collocation we also identify lemmas and \sm{Part-Of-Speech (POS)} information. 
\sm{We also keep track of the common nouns and proper nouns that are found in the sentence, that will play a central role in the steps described in the following subsections.}


After the NLP analysis, the text is segmented into sentences which are subsequently stored using a multi-field search engine. For each sentence, the following linguistic features are indexed as \sm{search  engine} fields:
\begin{itemize}
    \item Raw text (tokens)
    \item Collocations
    \item Lemmas (of collocations)
    \item POS (of collocations)
    \item Named entity type
\end{itemize}
\sm{Storing and indexing} the linguistic features allows us to use linguistic constrained \sm{search queries, that are crucial to improve the precision of the search results, as we will describe in the next subsection.}

%% file: 05_sentence_alignment.tex

\newcommand{\mincn}{min_{CN}}
\newcommand{\minpn}{min_{PN}}

\sm{In order to build a dataset composed of \textit{(sentence, paraphrase)} pairs, we define a group of reference \textit{sentences}, and, for each of them, we perform search engine queries to identify a set of candidate \textit{paraphrases}.}
\sm{Queries are based on the aforementioned linguistic features, which allows us} to constraint the resulting \sm{candidate paraphrases} to share a large number of information with the reference sentence used as search seed.
Moreover, the system can use information on the crawled article to which the reference sentence belongs to improve the precision of the search, such as the publication date and the article topic (paraphrases are expected to be found on articles with the same/similar publication dates).
\sm{As a result, we get a search system with very high precision and low recall, and the details of the whole sentence selection and alignment are reported in what follows.}

\subsubsection{Selection of the reference set}
\sm{Selecting the sentences that belong to the reference set \lr{is a procedure that is applied to all the crawled sentences}, and it depends on the number of detected proper nouns and common nouns. Our goal is to select sentences that are not too short and that involve the description of real-world facts.} 
In particular, given a sentence $s_i$, let us indicate with $CN(s_i)$ the set of common nouns in $s_i$ and with $PN(s_i)$ the set of proper nouns in $s_i$. 
The reference set $RS$ \sm{is defined as}
\begin{multline} \label{eq:1}
RS=\{rs_i : |CN(rs_i)| \geq \mincn \ \ \text{AND} \\ \ |PN(rs_i)| \geq \minpn\} \ ,
\end{multline}
where $rs_i$ is a sentence of the set and:
\begin{itemize}
    \item $\mincn$ is a system parameter indicating the minimum number of common nouns required in each sentence.
    \item $\minpn$ is a system parameter indicating the minimum number of proper nouns required  in each sentence.
\end{itemize}

\subsubsection{Constrained search (HCSSS)}
The sentences in the reference set are used as search seed  to retrieve candidate paraphrases. 
In particular, given a reference sentence $rs_i$, a search query \sm{is executed in order to retrieve} sentences $s_j$ that \sm{fulfill a number of constraints in terms of the common and proper nouns that they share with $rs_i$. 
In detail, we search for sentences $s_j$ such that:}
\begin{itemize}
    \item \sm{The proper nouns of $rs_i$ are included in $s_j$,} \begin{equation} \label{eq:2} PN(rs_i) \subseteq PN(s_j).\end{equation}
    \item \sm{There is a strong intersection between the common nouns in $rs_i$ and the ones in $s_j$,}
    \begin{eqnarray*}
    \hskip-2mm\text{If}\ |CN(rs_i)| > \mincn &\hskip-2mm\text{then} &\hskip-2mm\frac{| CN(rs_i) \cap CN(s_j) |}{|CN(rs_i)|} \geq \alpha \\\\
    \hskip-2mm\text{If}\ |CN(rs_i)| =\mincn &\hskip-2mm\text{then} &\hskip-2mm CN(rs_i) \subseteq CN(s_j),
    \end{eqnarray*}
\end{itemize}
where $\alpha \in [0,1]$ is a coverage parameter which filters out results having a small percentage of shared common nouns. \sm{We compactly report the function that checks for potential paraphrase pairs in Algorithm 1.} 
\begin{algorithm}
\SetAlgoNoLine
\LinesNumbered
\caption{Check if $s_j$ is compatible with a paraphrase of the reference sentence $rs_i$ ($[e]$ is the Boolean value of expression $e$).}
\SetKwFunction{FParaphrase}{isParaphrase}
\SetKwProg{Fn}{Function}{:}{}
\Fn{\FParaphrase{$rs_i$, $s_j$}}{
    \vspace{1mm}
    \KwData{reference sentence $rs_i$, candidate paraphrase $s_j$}
    \vspace{1mm}
    \KwResult{$true$ if $s_j$ is a paraphrase of $rs_i$, $false$ otherwise}
    \vspace{1mm}
    \If{$PN(rs_i) \subseteq PN(s_j)$}{
        \vspace{1mm}
        \If{$|CN(rs_i)| = \mincn$ } {
            \vspace{1mm}
            \Return $\left[CN(rs_i) \subseteq CN(s_j)\right]$
        }
        \vspace{1mm}
        \ElseIf{ $|CN(rs_i)| > \mincn$}{
            \vspace{1mm}
            \Return $\left[\frac{| CN(rs_i) \cap CN(s_j) |}{|CN(rs_i)|} \geq \alpha\right]$
        }
    }
    \sm{\Else{\Return $false$}}
}
\vspace{1mm}
\label{alg:find_para}
\end{algorithm}
\subsubsection{Filtering the candidate paraphrases}
The search procedure \sm{described so far} returns a large set of candidate paraphrases \sm{$\{para_1(rs_i)), \ldots ,  para_{n_i}(rs_i)\}$} for each reference sentence which has been provided as search seed $rs_i$, \sm{being $n_i$ the number of the returned candidates paraphrases}.
The list of results are ranked using the internal confidence score evaluated by the search engine which measures the degree of similarity between each result and the constrained query. 
The score usually varies in $[0,+\infty)$ \sm{and we can rescale it in $[0,1]$ by normalizing} with respect to the maximum score \sm{in the search results}. 
\sm{Those} results having a normalized score over a given threshold $\beta \in [0,1]$ are considered to be good paraphrases of the reference sentence $rs_i$ and they are selected for building a list of \sm{aligned} pairs $\{ (rs_i , para_1(rs_i)), ... , (rs_i , para_{\tilde{n}_i}(rs_i))\}$, \sm{being $\tilde{n}_i$ the number of sentences that are kept after the thresholding.}

\subsubsection{Building the final dataset pairs}

\lr{
At this stage, each of the pairs collected so far consists of a reference sentence and a candidate paraphrase $(rs_i , para_j(rs_i))$. 
The candidate sentence is constrained to include the same proper nouns of the reference sentence (see Eq. \ref{eq:2}) but in some cases it may also have additional proper nouns.
With the aim of ordering each pair with the same criterion, we ensure that the most informative sentence is in the first position of the pair. In our context, the informativeness of a sentence is measured by counting the number of proper nouns that it contains. Applying the procedure that is formalized in Algorithm 2, we get the final aligned dataset composed of what we generically refer to as \textit{(input, target)}.} 
\begin{algorithm}
\SetAlgoNoLine
\LinesNumbered
\caption{Build final dataset pairs.}
\SetKwFunction{FPair}{buildFinalPair}
\SetKwProg{Fn}{Function}{:}{}
\Fn{\FPair{$(rs_i , para_j(rs_i))$}}{
    \vspace{1mm}
    \KwData{aligned pair $(rs_i , para_j(rs_i))$}
    \vspace{1mm}
    \KwResult{final pair \textit{(input, target)}}
    \vspace{1mm}
    \If {$|PN(para_j(rs_i))| > |PN(rs_i)|$} {
        \vspace{1mm}
        \Return $(para_j(rs_i),rs_i)$
    }
    \vspace{1mm}
    \Else {
        \vspace{1mm}
        \Return $(rs_i,para_j(rs_i))$
    }
}
\vspace{1mm}
\label{alg:build_superivsed_pair}
\end{algorithm}

%% file: 06_case_study.tex
We applied the proposed method to the case of the Italian language, generating the dataset that we will use in the experiments of Section \ref{sec:exp}. The crawling procedure \lr{was performed by using a proprietary web monitoring system\footnote{http://www.mysnooper.net/} and it} \sm{downloaded} about $86000$ \sm{articles spanning over a time interval of about 2 months}, retrieved from Italian news and blog sites.
From these documents, we extracted more than $1$ million sentences that have been \sm{segmented and analyzed using a proprietary NLP platform\footnote{https://www.quest-it.com/natural-language-processing/}} \lr{developed by QuestIT\footnote{https://www.quest-it.com}. It is a pipeline-based processing platform consisting of more than $20$ layers of linguistic analysis and it uses a sliding-window SVM approach for POS Tagging, Lemmatization and Named Entity Recognition, while collocations are identified using a Search Tree combined with gazetteers.}
Sentences were indexed using Elastic Search\footnote{https://www.elastic.co/products/elasticsearch}. \sm{We selected} $\alpha=0.70$ and $\beta=0.70$, and the \sm{proposed method ($\mincn=\minpn=3$) identified} a reference set consisting of about $430000$ sentences, \sm{that yielded} a final paraphrasing dataset consisting of about $85000$ \sm{aligned} pairs in Italian language. 
\lr{Such final number of pairs is due to the fact that no paraphrases were identified for most of the reference sentences, since we imposed a high confidence score in the HCSSS procedure}.
In Figure \ref{grr} we report some examples of the \sm{aligned pairs \textit{(input, target)}} generated by the proposed method.

\begin{figure}
\noindent\fbox{\begin{minipage}{24em}
\footnotesize
\textbf{Input:}\texttt{ Brexit, no al secondo referendum: Parlamento boccia l'emendamento Brexit, May non si arrende: "Accordo entro 29 marzo alla nostra portata" Brexit, Parlamento affonda anche "No Deal": rinvio piu' vicino del divorzio Brexit, i Labour contro un secondo referendum.}
\\
\\
\textbf{Target:}\texttt{ Brexit, il Parlamento boccia l'accordo Brexit, il Parlamento vota no al “no deal” Brexit, no al secondo referendum: Parlamento boccia l'emendamento Brexit, i Labour contro un secondo referendum.}
\end{minipage}}
\vskip 1mm
\fbox{\begin{minipage}{24em}
\footnotesize
\textbf{Input:}\texttt{ Philippe Barbarin, 68 anni, cardinale e arcivescovo di Lione e uno dei maggiori prelati della Chiesa cattolica francese, e' stato condannato a 6 mesi di carcere con la condizionale: la sentenza e' stata pronunciata dal tribunale di Lione.}
\\
\\
\textbf{Target:}\texttt{ Il cardinale Philippe Barbarin, arcivescovo di Lione, condannato per aver coperto abusi sui minori, offre le sue dimissioni.}
\end{minipage}}
\vskip 1mm
\fbox{\begin{minipage}{24em}
\footnotesize
\textbf{Input:} \texttt{ La valutazione dell'agenzia di intelligence, nella quale i dirigenti affermano di avere un alto grado di fiducia, e' la piu' definitiva per ora tra quelle che legano il principe Bin Salman al delitto e complica gli sforzi dell'amministrazione Trump di salvare le relazioni con il suo stretto alleato in Medio Oriente.}
\\
\\
\textbf{Target:}\texttt{ La valutazione dell'agenzia di intelligence Usa e' la piu' autorevole per ora tra quelle che legano il principe Bin Salman al delitto e complica gli sforzi dell'amministrazione Trump di salvare le relazioni con il suo stretto alleato in Medio Oriente.}
\end{minipage}}
\vskip 1mm
\fbox{\begin{minipage}{24em}
\footnotesize
\textbf{Input:}\texttt{ Dorothy con le magiche scarpette rosse e tutti i suoi strampalati amici protagonisti dello spettacolo “Il mago di Oz”, il musical per bambini e famiglie in programma domenica 24 febbraio alle ore 16 al Teatro condominio di Gallarate.}
\\
\\
\textbf{Target:}\texttt{ Gallarate - Dorothy balla con le scarpette rosse nel musical “Il mago di Oz” - Bambini - Varese News.}
\end{minipage}}
\caption{Examples of aligned pairs automatically generated with the proposed method (Italian).}
\label{grr}
\end{figure}

%% file: 07_model.tex
Pointer networks \cite{vinyals2015pointer} are a special class of the so-called sequence-to-sequence models (i.e., models that process an input sequence and output another sequence \cite{sutskever2014sequence}), where the token to predict at each time step is obtained from a combination of two probability distributions. When generating the output sequence, the model either generates a word from the predefined vocabulary or it copies a word from the input sequence. This mechanism is helpful to catch and handle  Out-Of-Vocabulary (OOV) words, such as named entities, allowing the network to improve its performances in several NLP tasks (Machine Translation, Text Summarization, and others). 

We exploit Pointer networks to learn to generate paraphrases using the automatically created dataset of Section \ref{sec:data}. 
Given two sequences of words $x \in \mathcal{X}$, $y \in \mathcal{Y}$, that in the context of this paper are one the paraphrase of the other, the pointer network learns a mapping $f: \mathcal{X} \rightarrow \mathcal{Y}$.
We focus on a Pointer network model similar to the one of \cite{see2017get}, without any coverage mechanisms. In the rest of this section we will describe the two main computational blocks of the considered model, that are the plain sequence-to-sequence module that includes attention  (Section \ref{aa}) and the pointer-generator module (Section \ref{bb}).

\subsection{Sequence-to-Sequence Model with Attention}
\label{aa}
The Pointer network encodes the input sequence $x$ into a fixed-length internal representation using a Recurrent Neural Network (RNN). Then, another RNN generates the output sequence, following the classic sequence-to-sequence fashion. These networks are also referred to as encoder and decoder, respectively. 
We used LSTMs \cite{hochreiter1997long} for both the encoder (in particular, we used Bi-directional LSTMs (BiLSTMs)) and the decoder.
The encoder processes the whole sequence $x$ of length $N$, and its internal state at the end of the sequence, referred to as $h_{N}$, is used to initialize the state of the decoder. The encoder also returns all the intermediate states generated while processing the words of $x$, and computed using the classic recurrent scheme,
\begin{equation}
    h_i = \text{BiLSTM}(w_i, h_{i-1}) \ ,
\end{equation}
where BiLSTM is the function that represents the encoder and $w_i$ is the word embedding associated to the $i$-th word of the sequence. Similarly, the decoder includes its own internal state $s^t$ (where $t$ is the index associated to the words of the generated sequence -- we always use it as a superscript), obtained by feeding an LSTM with the previously generated word $w^{t-1}$ and the vector $\hat{h}^t$, that is computed by an attention layer \cite{bahdanau2014neural} over the encoder states. Formally,
\begin{eqnarray}
e_{(i)}^t = v_a' \cdot \text{tanh}(W_a\cdot[h_i, s^t]+ b_a)\\
\label{eq:attn_dist} a^t = \text{softmax}(e^t)\\ 
\hat{h}^t = \sum_{i=1}^{N} a_{(i)}^t \cdot h_i 
\end{eqnarray}
where $[,]$ is the concatenation operator and $a_{(i)}^t$ (resp. $e_{(i)}^t$) is the $i$-th element of vector $a^t$ (resp. $e^t$) of length $N$. The function tanh is the hyperbolic tangent activation, and the vectors $v_a, b_a$ and the matrix $W_a$ are parameters associated to the attention layer, and they are learned while training the  network.
The decoder state is then updated with
\begin{equation}
    s^t = \text{LSTM}([w^{t-1}, \hat{h}^t], s^{t-1}) \ .
\end{equation}
Finally, a projection layer transforms the state $s_t$, concatenated with $\hat{h}^t$, into a probability distribution on the words in the vocabulary
\begin{equation}
    P_{\text{vocab}}^t = \text{softmax}(V\cdot[s^t, \hat{h}^t]+ b)\label{eq:p_voc},
\end{equation}
where the matrix $V$ and vector $b$ collect weights that have to be learnt. We will use the notation $P_{\text{vocab}}^t(\texttt{w})$ to indicate the probability of generating word $\texttt{w}$ at time $t$.

\subsection{Pointer Generator}
\label{bb}
What we described so far is a plain generator that selects the next word to produce at each time instant, i.e., the one that is most likely accordingly to Eq. \ref{eq:p_voc}.
When introducing a pointer-based mechanism, the generator considers a combination between the probability of emitting a word from the fixed vocabulary (Eq. \ref{eq:p_voc}) and the probability of emitting (copying) a word taken from the input sequence $x$. The latter probability, indicated with $P^t_{\text{copy}}$, is computed by exploiting the attention scores $a^t$ of Eq. \ref{eq:attn_dist}. In detail,
\begin{equation}
    P_{\text{copy}}^t(\texttt{w}) = \sum_{i: \texttt{w}_i = \texttt{w}}a_{(i)}^t \ ,
\end{equation}
where the summation is only needed in the case in which a word is repeated multiple times in the input sentence, and all the probabilities $a_i^t$ associated to the instances of such word must be accumulated. In order to combine $P_{\text{vocab}}^t(\texttt{w})$ and $P_{\text{copy}}^t(\texttt{w})$ into the final generation probability $P^t(\texttt{w})$, a weighting score $p_{\text{gen}}^t \in [0,1]$ is introduced, that depends on the decoder state $s^t$, the context vector $\hat{h}^t$ and the embedding of the previously generated word $w^{t-1}$,
\begin{equation}
    p_{\text{gen}}^t = \sigma(W_{\text{gen}}\cdot[w^{t-1}, s^t, \hat{h}^t]+b_{\text{gen}}).
\end{equation}
The parameters $W_{gen}$ and $b_{gen}$ are learned jointly with the whole networks, and $\sigma$ is the sigmoid activation function.
Qualitatively, the model can learn when it should copy from the source input and when it is not the case. 
The final probability of generating a word $\texttt{w}$ at time $t$ is then
\begin{equation}
    P^t(\texttt{w}) = p_{\text{gen}}^t\cdot P^t_{\text{vocab}}(\texttt{w}) + (1- p_{\text{gen}}^t)\cdot P^t_{\text{copy}}(\texttt{w}) \ .
\end{equation}

It is worth noticing that $\texttt{w}$ may be an OOV word. In such  case $P^t_{\text{vocab}}(\texttt{w})$ would be zero. Similarly, if $\texttt{w}$ does not belong to the input sentence, $P^t_{\text{copy}}(\texttt{w})$ = 0.

%% file: 08_experimental_results.tex
Given the dataset created with the procedure of Section  \ref{sec:data}, we evaluate the performances of different sequence-to-sequence models in the task of paraphrasing. Results are compared by means of the popular ROUGE score \cite{lin2004rouge}, with the aim of showing the feasibility of the proposed approach. However, we also present some qualitative results to emphasize the limits of the compared models.

In detail, we compared a plain sequence-to-sequence model with attention (Section \ref{aa}) and the Pointer network that we obtain once we add a pointer-based generator (Section \ref{bb}). We also evaluate the case in which we pretrain a part of the Pointer network in an autoencoding task, where the network is asked to simply generate the input sentence. During such pretraining, both the attention and the poitining mechanisms were turned off, to avoid the network to simply learn the trivial solution of always copying the input text.
We exploited the PAISA' dataset \cite{lyding2014paisa} in the pretraining stage, that is a large corpus of Italian texts taken from the web. 

We used the same architecture of \cite{see2017get}, changing the size of the vocabulary to $30k$ words, and the maximum length of input and output sequences to $70$ tokens. Our paraphrasing corpus was divided into two disjoint subsets of about $72k$ and $12k$ pairs, used as training and validation sets, respectively.
Furthermore, we considered a test set of $1384$ pairs (excluded from the training and validation sets).
Learning was stopped after $15$ consecutive epochs without any improvements on the validation set. All models were optimized with Adam \cite{kingma2014adam}, with learning rate $0.001$ and clipping the gradient norm to $4.0$. 
During natural language generation, a beam search with a beam size of $10$ is exploited.

Our first experiences led to not promising results (that we avoid reporting). We found that the networks were badly biased by the presence of multiple paraphrases of the same sentence. After filtering the data to ensure that for each input sentence there was only one target paraphrase, and after having augmented the data including also flipped pairs, we ended up with a training, validation, test set of $35k$, $10k$ and $1485$ examples, respectively. 

The results we obtained using such data are reported in Table \ref{exp:results}. 
\begin{table}[!ht]
\caption{ROUGE-$1$,$2$,$l$ $F_1$ score on -- test data. }\label{exp:results}
\centering
\begin{tabular}{l|ccc}
\hline
 & \multicolumn{3}{c}{ROUGE}   \\
  &    $1$   &  $2$    &    $l$   \\ \hline
 \textsc{sequence-to-sequence + attention} &  44.39     &    31.71  &  41.36  \\
 \textsc{pointer network}&       \textbf{60.66}     &  \textbf{50.21}     & \textbf{56.64}  \\
 \textsc{pretrained pointer network} &   60.51    & 49.88   & 56.29   \\ \hline
\end{tabular}
\end{table}
We evaluated the performances measuring different ROUGE-based metrics, that are the F1 scores in the case of ROUGE-1, ROUGE-2 and ROUGE-$l$ (see \cite{lin2004rouge}). These metrics quantify the word-overlap, bigram-overlap, and longest common sequence between the target text and the predicted paraphrase, respectively. 
The results we obtained are in-line with the ones commonly obtained in Text Simplification and Single-sentence Summarization \cite{zhang2017semantic}, which are tasks related to the one we consider. They suggest that the proposed dataset construction procedure is a feasible way to collect data to train Pointer networks for paraphrasing.
The copying mechanism (Pointer network) significantly improves the ROUGE scores with respect to plain sequence-to-sequence, while the pre-training stage does not introduce further improvements. While the former results was expected, the latter might due to the fact that the pretraining stage did not use any pointers and/or attention (for the reasons we described earlier), making it hard to transfer the pretrained model toward the final Pointer network.

An aspect that might require improvements to the Pointer network is that all the trained models are strongly biased toward the reproduction of the exact input sentence. As a matter of fact, the input sentence and the target paraphrase are sometimes very similar, due to the intrinsic property of the paraphrasing task. This behaviour is evident when looking at the generations reported in in the example of Figure \ref{uff}.
We leave this point open for further investigation on improved neural models for paraphrasing.
\begin{figure}
\fbox{\begin{minipage}{24em}
\footnotesize
\textbf{Input:}\texttt{ – tutto il programma velate, varese – domenica 23 a velate un'ospite illustre: ada cattaneo vestir\'a i panni della regina delle nevi per raccontare le incantevoli tradizioni di questo periodo dell'anno.}
\\
\\
\textbf{Target:}\texttt{ in arrivo la prossima domenica 23 a velate un'ospite illustre: ada cattaneo vestir\'a i panni della regina delle nevi per raccontare le incantevoli tradizioni di questo periodo dell'anno.}
\\
\\
\textbf{Seq-to-seq:}\texttt{ tutto il programma 2019, varese – domenica 23 a novembre un'ospite illustre : gino cattaneo, i panni della regina delle nevi per raccontare le <UNK> tradizioni di questo periodo dell'anno.}
\\
\\
\textbf{Pointer-net:}\texttt{ varese, domenica 23 a velate un'ospite illustre: ada cattaneo vestir\'a i panni della regina delle nevi per raccontare le incantevoli tradizioni di questo periodo dell'anno.}
\end{minipage}}
\caption{Example of a common issue in the considered neural models. Generated sentences are sometimes too similar to the input sentence.}
\label{uff}
\end{figure}

%% file: 09_conclusion_future_works.tex

In this paper \sm{we presented} a method for \sm{the} automatic generation of \sm{aligned} corpora \sm{that can be used  to train machine learning-based} paraphrasing systems.
Using massive/focused crawling, a large number of web articles are downloaded from news and blog websites.
Each document is processed using Natural Language Processing (NLP) techniques, and their contents are indexed using a search engine (multi-field inverse index) which includes the set of linguistic features extracted by the NLP analysis. 
A selection procedure identifies a reference set of sentences \sm{and, for each of them}, a linguistic\sm{ally} constrained search query is submitted to the search engine, \sm{retrieving} a list of results that \sm{are marked as paraphrases of the considered sentence}.
The \sm{proposed} method has been used to produce a  dataset of \sm{paraphrases in the case of the} Italian language, crawling more than $86000$ articles from Italian news and blog sites for $2$ months, leading to $1$ million of sentences \sm{indexed by the search engine}. 
\sm{After the constrained search procedure, we ended up with a dataset of $85000$ aligned pairs.}
The \sm{quality} of the dataset has been \sm{evaluted} by \sm{exploiting} it to train a neural \sm{paraphrasing} model \sm{with} a sequence-to-sequence architecture with pointer-based generator.

The \sm{data generation} procedure is continuously working, and new articles are downloaded from \sm{multiple} web sources, increasing the size of the dataset. \sm{We are currently working on applying the proposed method to setup a dataset of English paraphrases. We are also considering the possibility of injecting more knowledge into the paraphrase generation in the framework of Learning from Constraints \cite{gori2013constraint,gnecco2014learning,gnecco2014theoretical,melacci2012unsupervised}, and focusing in an online learning setting \cite{frandina2013variational}.}